\pdfoutput=1
\documentclass[letterpaper,10pt,conference]{IEEEtran}

\usepackage[T1]{fontenc}

\usepackage{newtxtext,newtxmath}

\usepackage{graphicx}
\usepackage{amsmath}
\usepackage{booktabs}
\usepackage{placeins}
\usepackage{array}
\usepackage{multirow}
\usepackage{tabularx}

\usepackage{xurl}

\usepackage[hidelinks]{hyperref}

\title{Systematic Evaluation of Depth Backbones and Semantic Cues for Monocular Pseudo-LiDAR 3D Detection}

\author{
\IEEEauthorblockN{Samson Oseiwe Ajadalu}
\IEEEauthorblockA{University of Toronto\\
\texttt{s.ajadalu@mail.utoronto.ca}}
}

\begin{document}
\maketitle

\begin{abstract}
Monocular 3D object detection offers a low-cost alternative to LiDAR, yet remains less accurate due to the difficulty of estimating metric depth from a single image. We systematically evaluate how depth backbones and feature engineering affect a monocular Pseudo-LiDAR pipeline on the KITTI validation split. Specifically, we compare NeWCRFs (supervised metric depth) against Depth Anything V2 Metric-Outdoor (Base) under an identical pseudo-LiDAR generation and PointRCNN detection protocol. NeWCRFs yields stronger downstream 3D detection, achieving 10.50\% AP$_{3D}$ at IoU$=0.7$ on the Moderate split using grayscale intensity (Exp~2). We further test point-cloud augmentations using appearance cues (grayscale intensity) and semantic cues (instance segmentation confidence). Contrary to the expectation that semantics would substantially close the gap, these features provide only marginal gains, and mask-based sampling can degrade performance by removing contextual geometry. Finally, we report a depth-accuracy-versus-distance diagnostic using ground-truth 2D boxes (including Ped/Cyc), highlighting that coarse depth correctness does not fully predict strict 3D IoU. Overall, under an off-the-shelf LiDAR detector, depth-backbone choice and geometric fidelity dominate performance, outweighing secondary feature injection.
\end{abstract}

\begin{figure*}[t]
\centering
\includegraphics[width=\linewidth]{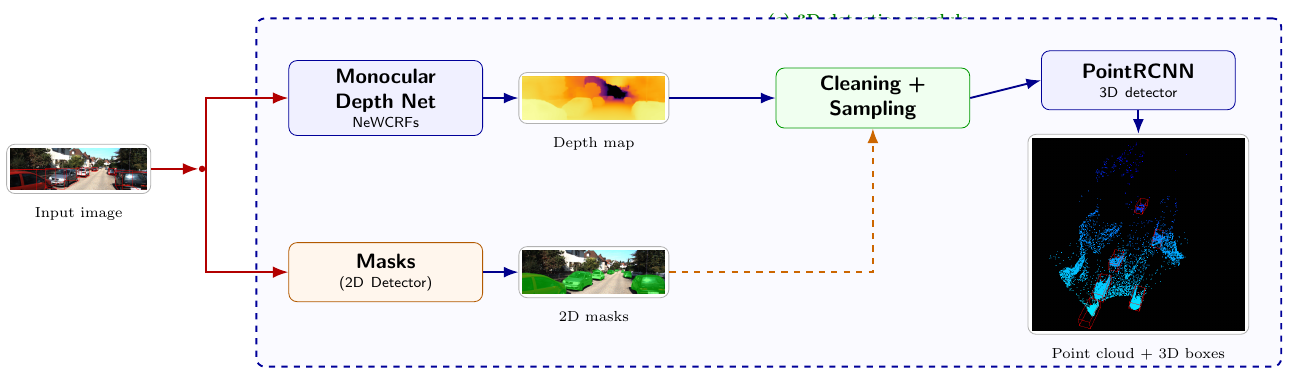}
\caption{Overall monocular pseudo-LiDAR pipeline.}
\label{fig:pipeline}
\end{figure*}

\section{Introduction}
Monocular 3D object detection is a crucial yet challenging task for autonomous driving and robotics.
Cameras are inexpensive and provide dense appearance information, but without stereo or LiDAR, recovering accurate metric depth from a single image remains difficult.
On the KITTI benchmark~\cite{Geiger2012CVPR}, this limitation historically creates a large performance gap between monocular and LiDAR-based 3D detectors~\cite{chen2020cvpr}.

A key step toward closing this gap was the Pseudo-LiDAR paradigm~\cite{Wang_2019_CVPR}, which converts an estimated depth map into a 3D point cloud and then applies a LiDAR detector.
This showed that the \emph{3D representation} (point cloud versus image-plane reasoning) can be a dominant factor for downstream detection performance.
Although strong depth networks improve monocular Pseudo-LiDAR, monocular accuracy still trails stereo and LiDAR settings, especially at longer ranges and under occlusion.

Recent work has also explored injecting semantic cues from RGB into pseudo-LiDAR to complement geometry.
For example, instance-guided pseudo-LiDAR representations append mask-related signals to the point cloud~\cite{Gao20242DIP}.
However, it remains unclear how much such feature engineering helps when using standard LiDAR detectors whose input channels and inductive biases were designed for physical LiDAR measurements.

In this work, we perform a systematic evaluation of depth backbones and semantic feature variants within a monocular Pseudo-LiDAR pipeline on KITTI.
We study two questions:
(1) How does a supervised metric depth model compare to a foundation metric depth model in terms of downstream 3D detection under an identical detector and training protocol?
(2) What is the relative impact of engineered semantic features (intensity channels, mask confidence, and mask-based sampling) compared to geometry alone?
We evaluate NeWCRFs~\cite{Yuan_2022_CVPR} and Depth Anything V2 Metric-Outdoor (Base)~\cite{depth_anything_v2}, and feed the resulting pseudo-point clouds into PointRCNN~\cite{Shi_2019_CVPR} using OpenPCDet~\cite{openpcdet2020}.

Our ablations isolate three factors:
\begin{itemize}
    \item \textbf{Depth backbone effects:} comparing NeWCRFs against Depth Anything V2 Metric-Outdoor (Base) under the same pseudo-LiDAR feature setting and detector protocol.
    \item \textbf{Semantic feature channels:} injecting grayscale intensity and instance segmentation confidence into the pseudo-LiDAR points.
    \item \textbf{Context manipulation:} mask-based point selection that increases foreground density but reduces background/ground structure.
\end{itemize}

Across experiments, we observe that \textbf{geometry dominates performance} in this pipeline.
Within NeWCRFs-based variants, grayscale intensity (Exp~2) achieves the strongest performance (10.50\% AP$_{3D}$ at IoU$=0.7$ on Moderate), while a zero-intensity control (Exp~7) remains close, indicating limited dependence on the 4th channel.
Mask-confidence injection (Exp~4) can preserve BEV overlap but degrades strict 3D IoU, and mask-only point selection (Exp~5) can hurt by removing contextual geometry needed for stable 3D reasoning.
When swapping only the depth backbone, NeWCRFs remains stronger than Depth Anything V2 Metric-Outdoor under the same detector/training setup, especially under the strict IoU$=0.7$ criterion.

\section{Related Work}
\label{sec:related}

\subsection{Monocular 3D Object Detection and Pseudo-LiDAR}
Early monocular 3D object detection methods often relied on geometric assumptions and learned priors,
rather than accurate metric depth. For example, Mono3D~\cite{Chen_2016_CVPR} combines 2D detection cues with
ground-plane constraints and shape priors to score candidate 3D boxes, but such approaches remain far from
LiDAR-level performance on autonomous driving benchmarks.

The Pseudo-LiDAR paradigm~\cite{Wang_2019_CVPR} changed this landscape by explicitly reconstructing a 3D point
cloud from image-based depth and then applying mature LiDAR-based detectors.
This representation enables monocular pipelines to reuse LiDAR detection backbones and can yield large gains
compared to purely image-plane formulations. Related monocular variants also explored pseudo-LiDAR point clouds
for 3D detection in driving settings~\cite{Weng_2019_ICCV_Workshops}. Using stronger monocular depth estimation
(e.g., DORN~\cite{FuCVPR18-DORN}) further improves performance, though monocular settings still trail stereo and LiDAR,
especially at longer ranges.

Follow-up work targeted depth accuracy and robustness. Pseudo-LiDAR++~\cite{you2020pseudolidar} improves depth
quality for far-away objects and reports strong gains for stereo-based 3D detection on KITTI.
In parallel, several end-to-end monocular detectors were proposed to avoid explicit pseudo-point-cloud conversion.
MonoGRNet~\cite{Qin2019MonoGRNet} introduces explicit geometric reasoning for 3D localization, while D$^4$LCN~\cite{Ding_2020_CVPR}
uses depth-guided dynamic convolutions to better align image features with 3D structure.
More recently, DD3D~\cite{park2021dd3d} explores single-stage monocular 3D detection that benefits from depth pretraining
without requiring pseudo-LiDAR at inference.

A closely related approach to handling depth uncertainty is CaDDN~\cite{CaDDN}, which projects image features into 3D
using \emph{categorical depth distributions} rather than a single deterministic depth value.
This probabilistic projection explicitly models depth ambiguity and can improve robustness to noisy depth.
Our work complements this line from a different angle. Instead of modifying the detector to handle uncertainty,
we study how far a standard pseudo-LiDAR pipeline can go when \emph{input depth quality} is improved, and how much additional
gain (if any) comes from injecting semantic features into the point cloud.

\subsection{Depth Estimation Models for Pseudo-LiDAR}
Progress in monocular depth estimation directly benefits pseudo-LiDAR pipelines.
NeWCRFs~\cite{Yuan_2022_CVPR} is a strong supervised depth model that refines depth using neural window fully-connected CRFs,
producing high-quality depth maps on driving scenes.
In contrast, Depth Anything~\cite{depth_anything_v1,depth_anything_v2} represents foundation-model style depth estimators trained on
large and diverse data, offering strong generalization. The family includes both relative-depth models and released
metric fine-tuned variants for outdoor scenes, which are suitable when absolute scale is required.
In our experiments, we compare a domain-specialized model (NeWCRFs) against a fine-tuned generalist model family (Depth Anything)
to quantify the impact of depth fidelity on downstream 3D detection.

\subsection{Semantic Feature Augmentation for 3D Detection}
Beyond geometry, semantic cues from RGB may help monocular 3D detection when integrated effectively.
Gao \textit{et al.}~\cite{Gao20242DIP} propose instance-guided pseudo-LiDAR by appending a mask-confidence channel to each 3D point and
modifying the network to exploit this additional information.
More broadly, the idea of ``painting'' point clouds with semantic outputs from images has proven useful in sensor fusion.
PointPainting~\cite{Vora_2020_CVPR} projects semantic segmentation scores onto LiDAR points and shows improved 3D detection.
Motivated by these ideas, our work evaluates whether simple semantic feature injection (e.g., grayscale intensity or mask confidence)
improves pseudo-LiDAR detection under strong depth backbones when using a largely standard LiDAR detector architecture.

\begin{figure*}[t]
\centering
\includegraphics[width=\linewidth]{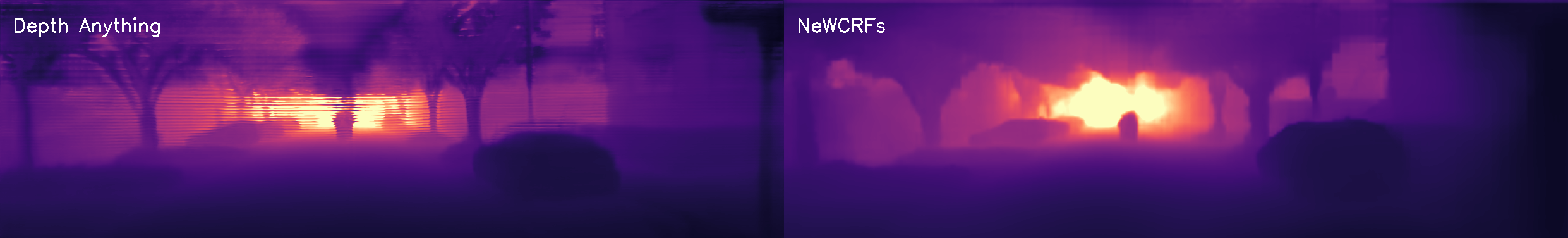}
\caption{Qualitative depth comparison (same KITTI frame): Depth Anything V2 Metric-Outdoor shows banding/over-smoothing relative to NeWCRFs.}
\label{fig:depth_newcrfs_da}
\end{figure*}

\section{Methodology}
\label{sec:method}

Figure~\ref{fig:pipeline} summarizes our monocular 3D detection pipeline.
It consists of three stages: (1) depth estimation, (2) pseudo-LiDAR generation, and (3) 3D object detection.
We define a set of experimental variants (Section~\ref{sec:exp_variants}) to isolate the effects of
depth quality and semantic feature engineering.

\subsection{Depth Estimation}
Given a single monocular RGB image from the KITTI front camera, we predict a dense depth map $D(u,v)$ at each pixel $(u,v)$.
We compare two monocular depth backbones in the full pseudo-LiDAR detection pipeline:
\begin{itemize}
    \item \textbf{NeWCRFs}~\cite{Yuan_2022_CVPR}: a supervised monocular depth estimator trained to predict metric depth for driving scenes.
    \item \textbf{Depth Anything V2 Metric-Outdoor (Base)}~\cite{depth_anything_v2}: metric foundation depth model; we use the public Metric-Outdoor Base checkpoint for pseudo-LiDAR generation.

\end{itemize}

\subsection{Pseudo-LiDAR Point Cloud Generation}
We convert the predicted depth map into a 3D point cloud following the pseudo-LiDAR formulation~\cite{Wang_2019_CVPR}.
For each pixel $(u,v)$ with depth $Z_c = D(u,v)$, we unproject into 3D using the KITTI camera intrinsics:
\begin{equation}
X = (u - c_x)\frac{Z_c}{f_x}, \quad
Y = (v - c_y)\frac{Z_c}{f_y}, \quad
Z = Z_c,
\end{equation}
where $(f_x,f_y)$ are focal lengths and $(c_x,c_y)$ is the principal point.
This yields 3D points $(X,Y,Z)$ in the rectified camera coordinate frame (meters). We then transform these points
to the KITTI Velodyne (LiDAR) coordinate frame using the provided calibration (rectification $R_{0}$ and extrinsics
$T_{\text{velo}\rightarrow\text{cam}}$) before training and inference with PointRCNN.
We retain points within a driving-relevant region of interest (ROI) by filtering invalid depth values.

\paragraph{Intensity / Feature Channel}
LiDAR detectors typically consume 4D point inputs $(x,y,z,I)$, where $I$ is reflectance.
To avoid architectural changes, we always provide a 4th channel and populate it with different values
(e.g., grayscale intensity or mask confidence) depending on the experimental configuration.

\paragraph{Instance Mask Probabilities}
For experiments that require semantic cues (Exp~4 and Exp~5), we obtain a per-pixel instance segmentation probability
for the \textit{car} class using Mask R-CNN~\cite{8237584}.
We initialize Mask R-CNN with COCO-pretrained weights and fine-tune it on the KINS dataset~\cite{qi2019amodal},
which provides instance mask annotations aligned with the KITTI object detection images.
We fine-tune on the KINS training split (3,712 frames) and apply the model to generate mask probabilities for the KITTI images (evaluated on the validation split of 3,769 frames). We convert instance masks into a per-pixel car confidence map in $[0,1]$ by assigning pixels inside each car mask the instance score (background $0$; overlaps take the max), and use this map either as the 4th channel or for mask-guided sampling.

\subsection{3D Object Detection with PointRCNN}
We detect 3D bounding boxes from the pseudo-LiDAR point cloud using PointRCNN~\cite{Shi_2019_CVPR}
implemented in OpenPCDet~\cite{openpcdet2020}.
PointRCNN is a two-stage detector consisting of (i) point-wise foreground segmentation and proposal generation,
and (ii) a refinement stage that pools point features within each proposal to regress the final 3D box.

For each experimental configuration, we train a separate PointRCNN model from scratch using the corresponding
pseudo-LiDAR inputs. We keep the detector architecture, data augmentation, and hyperparameters consistent
across experiments to isolate the effects of the input data (depth backbone and semantic features).
\begin{table}[t]
\centering
\caption{Summary of pseudo-LiDAR variants (Car). Same PointRCNN protocol; only pseudo-LiDAR construction changes. Exp~0 uses GT 2D boxes.}
\label{tab:variant_summary}

\scriptsize
\setlength{\tabcolsep}{2.2pt}
\renewcommand{\arraystretch}{1.05}

\begin{tabularx}{\columnwidth}{@{}p{0.22\columnwidth}X@{}}
\toprule
Variant & Key change \\
\midrule
Exp~0 & Heuristic box fit (priors + PCA yaw) using GT 2D boxes; cluster frustum points and fit box. \\
Exp~2 & NeWCRFs, full-scene pseudo-LiDAR; $I$ = grayscale. \\
Exp~4 & NeWCRFs, full-scene pseudo-LiDAR; $I$ = mask confidence. \\
Exp~5 & NeWCRFs, mask-guided sampling; reduced context; 40k points. \\
Exp~7 & NeWCRFs, full-scene pseudo-LiDAR; $I=0$. \\
DA Metric (Exp~2) & Depth backbone swapped to Depth Anything V2 Metric-Outdoor (Base); same construction as Exp~2. \\
\bottomrule
\end{tabularx}
\end{table}
\begin{table*}[!t]
\centering
\caption{KITTI validation results (Car, AP$_{40}$).}
\label{tab:car_apbev_ap3d}

\scriptsize
\setlength{\tabcolsep}{2.8pt}
\renewcommand{\arraystretch}{1.05}

\begin{tabular*}{\textwidth}{@{\extracolsep{\fill}} l ccc ccc}
\toprule
& \multicolumn{3}{c}{AP$_{BEV}$/AP$_{3D}$ (\%), IoU=0.5} & \multicolumn{3}{c}{AP$_{BEV}$/AP$_{3D}$ (\%), IoU=0.7} \\
\cmidrule(lr){2-4}\cmidrule(lr){5-7}
Experiment (features) & Easy & Moderate & Hard & Easy & Moderate & Hard \\
\midrule
Exp~0: Heuristic Box Fitting (Priors + PCA Yaw) & 4.40/2.56 & 4.22/2.17 & 4.28/2.35 & 0.94/0.22 & 1.07/0.28 & 0.64/0.29 \\
Exp~2: NeWCRFs + grayscale            & \textbf{29.72/26.90} & \textbf{22.60/19.06} & \textbf{19.36/17.97} & \textbf{13.85/11.88} & \textbf{11.72/10.50} & \textbf{11.20/9.98} \\
Exp~4: NeWCRFs + mask confidence      & 28.43/26.02 & 22.17/18.45 & 18.30/16.93 & 13.41/4.01  & 11.36/5.73 & 10.89/5.58 \\
Exp~5: NeWCRFs + mask-guided sampling     & 26.68/18.65 & 18.87/17.52 & 17.00/15.90 & 5.97/3.88   & 10.97/9.81 & 10.12/9.89 \\
Exp~7: NeWCRFs + zero intensity       & 28.38/25.46 & 19.77/18.53 & 18.53/16.81 & 13.25/10.90 & 11.49/9.83 & 10.81/9.90 \\
\bottomrule
\end{tabular*}
\end{table*}

\begin{table*}[!t]
\centering
\caption{Depth backbone comparison under the same detector/training protocol (Car, KITTI validation, AP$_{40}$).
Values are AP$_{BEV}$/AP$_{3D}$ (\%). Best in each difficulty column (based on AP$_{3D}$) is bolded separately for IoU=0.5 and IoU=0.7.}
\label{tab:depth_model_comp}

\scriptsize
\setlength{\tabcolsep}{2.6pt}
\renewcommand{\arraystretch}{1.05}

\begin{tabular*}{\textwidth}{@{\extracolsep{\fill}} l ccc ccc}
\toprule
& \multicolumn{3}{c}{AP$_{BEV}$/AP$_{3D}$ (\%), IoU=0.5} & \multicolumn{3}{c}{AP$_{BEV}$/AP$_{3D}$ (\%), IoU=0.7} \\
\cmidrule(lr){2-4}\cmidrule(lr){5-7}
Depth Model & Easy & Moderate & Hard & Easy & Moderate & Hard \\
\midrule
NeWCRFs~\cite{Yuan_2022_CVPR}
  & \textbf{29.72/26.90} & \textbf{22.60/19.06} & \textbf{19.36/17.97}
  & \textbf{13.85/11.88} & \textbf{11.72/10.50} & \textbf{11.20/9.98} \\
Depth Anything V2 Metric-Outdoor (Base)~\cite{depth_anything_v2}
  & 25.59/22.00 & 19.36/17.85 & 17.68/16.33
  & 11.81/10.53 & 11.15/9.79 & 10.32/9.87 \\
\bottomrule
\end{tabular*}
\end{table*}
\subsection{Experimental Variants}
\label{sec:exp_variants}

We study how (i) the depth backbone and (ii) simple per-point feature choices affect monocular pseudo-LiDAR detection, while keeping the detector (PointRCNN) and training protocol fixed. Unless otherwise stated, variants use NeWCRFs depth. We additionally report a non-learning baseline (Exp~0) using oracle (ground-truth) 2D boxes to factor out 2D detection errors.
Table~\ref{tab:variant_summary} summarizes the pseudo-LiDAR variants evaluated in this work.

\begin{itemize}
    \item \textbf{Depth backbone:} NeWCRFs vs.\ Depth Anything V2 Metric-Outdoor (Base) under the same pseudo-LiDAR construction (Exp~2 setting), isolating geometry differences.
    \item \textbf{4th channel (intensity) ablation:} grayscale intensity (Exp~2) vs.\ mask-confidence intensity (Exp~4) vs.\ a zero-intensity control (Exp~7).
    \item \textbf{Context preservation:} full-scene pseudo-LiDAR (Exp~2/Exp~4/Exp~7) vs.\ mask-guided point selection that reduces background/ground context (Exp~5).
    \item \textbf{Non-learning baseline (Exp~0):} heuristic frustum-based box fitting using \emph{oracle (GT) 2D boxes}, clustering, PCA/SVD yaw, and class size priors, providing a geometry-only reference before learned 3D detection.
\end{itemize}



\section{Experiments and Results}
\label{sec:results}
\subsection{Experimental Setup}
\label{sec:exp_setup}
Our 3D detector is PointRCNN~\cite{Shi_2019_CVPR} implemented in OpenPCDet~\cite{openpcdet2020}. We keep the
PointRCNN architecture and optimization schedule fixed across ablations to isolate the impact of the
pseudo-LiDAR input (depth backbone and 4th-channel feature variants). Unless otherwise stated, pseudo-LiDAR
points are generated by back-projecting dense depth into 3D using KITTI intrinsics, transforming from camera
to Velodyne coordinates using KITTI calibration, and retaining only valid depth values in the range $1.0 < D(u,v) < 60.0$\,m before back-projection. For NeWCRFs-based variants, we use the predicted depth directly before back-projection.
The pseudo-LiDAR intensity channel $I$ is normalized to $[0,1]$ (grayscale$/255$ or a confidence value,
depending on the experiment). For Exp~7, we generate a geometry-only pseudo-LiDAR by setting the intensity
channel to zero ($I=0$) while keeping the same 3D points and preprocessing as Exp~2. For Exp~5, we use a denser
point budget (40{,}000 points) and update the OpenPCDet point sampling (\texttt{sample\_points}) accordingly
to preserve this increased density during training and evaluation.

Table~\ref{tab:exp_setup} summarizes the common training and evaluation settings shared by all experiments.
\begin{figure*}[t]
\centering
\includegraphics[width=\linewidth]{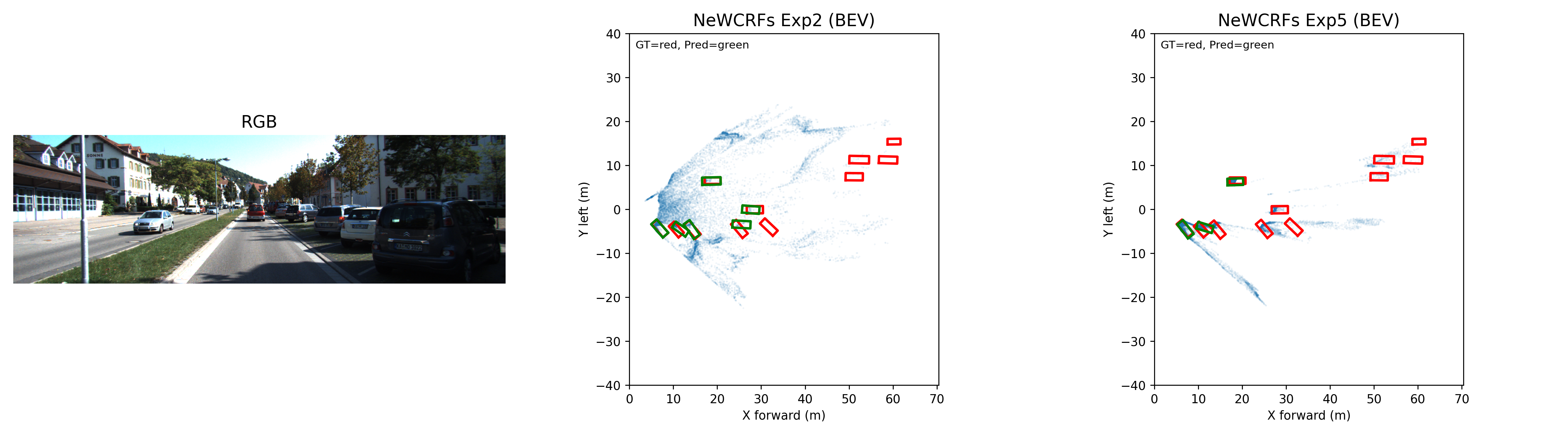}
\caption{\textbf{Qualitative BEV comparison: Exp~2 vs Exp~5 (NeWCRFs).} Blue points show the pseudo-LiDAR point cloud. Red boxes are KITTI ground-truth 3D boxes and green boxes are PointRCNN predictions (BEV). Exp~5 uses mask-guided point selection that preserves dense car points but strongly reduces background/road context compared to Exp~2 (full-scene cloud), which can destabilize localization and increase errors at high IoU.}
\label{fig:bev_exp2_exp5}
\end{figure*}
\begin{figure*}[t]
\centering
\includegraphics[width=\linewidth]{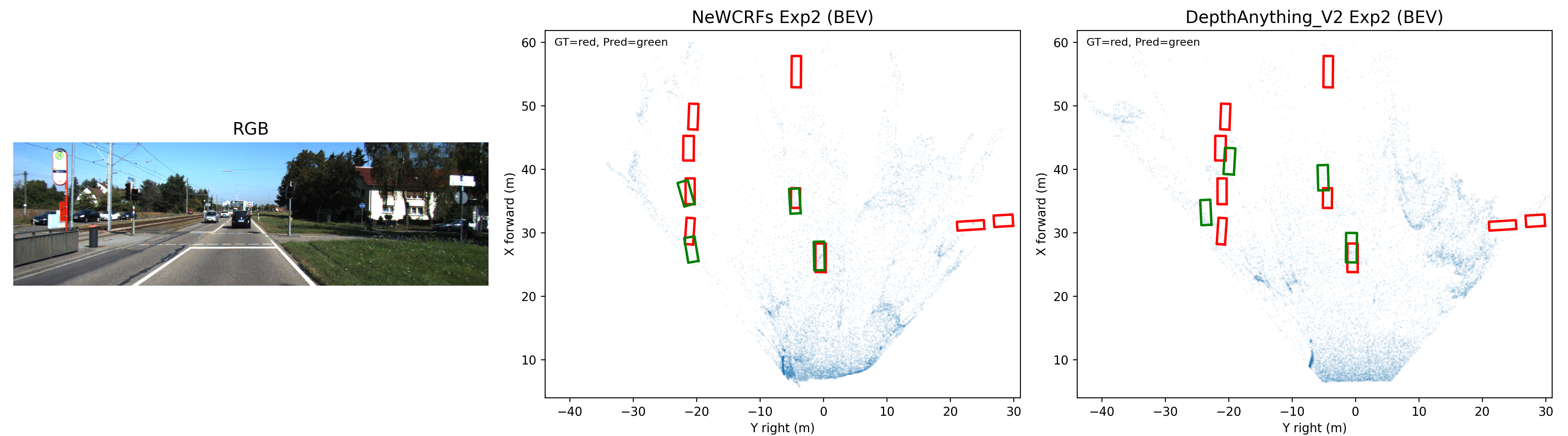}
\caption{\textbf{Qualitative BEV comparison (Exp~2 setting):} NeWCRFs vs.\ Depth Anything V2 Metric-Outdoor under the same PointRCNN protocol.}
\label{fig:bev_newcrfs_da}
\end{figure*}
\begin{table}[t]
\centering
\caption{Common experimental setup (Car, KITTI validation).}
\label{tab:exp_setup}

\scriptsize
\setlength{\tabcolsep}{2.2pt}
\renewcommand{\arraystretch}{1.05}

\begin{tabularx}{\columnwidth}{@{}p{0.28\columnwidth}X@{}}
\toprule
Item & Setting \\
\midrule
Split &
KITTI train split (3,712 frames) for training; results on the validation split (3,769 frames). \\
Detector &
PointRCNN~\cite{Shi_2019_CVPR} (OpenPCDet~\cite{openpcdet2020}); trained separately per variant. \\
Training (fixed) &
80 epochs; batch size 2/GPU; \texttt{adam\_onecycle}; LR $0.01$; weight decay $0.01$; grad clip 10. \\
Input &
Pseudo-LiDAR points $(x,y,z,I)$ in Velodyne coordinates. \\
Depth filter & Keep $1 < D(u,v) < 60$\,m before back-projection. \\
Point sampling &
\texttt{NUM\_POINTS}: 16,384 (Exp~2/4/7) and 40,000 (Exp~5). \\
Range filter &
\texttt{mask\_points\_and\_boxes\_outside\_range} enabled (training). \\
Metric &
AP$_{BEV}$/AP$_{3D}$ (AP$_{40}$) at IoU $0.5$ and $0.7$ (Easy/Mod/Hard). \\
\bottomrule
\end{tabularx}
\end{table}

\subsection{3D Detection Performance (Car)}
Within the NeWCRFs-based experimental variants (Table~\ref{tab:car_apbev_ap3d}), Exp~2 (grayscale intensity)
achieves the strongest overall performance. Exp~7 (zero intensity) remains close, indicating that PointRCNN relies
primarily on geometric structure in pseudo-LiDAR and only weakly benefits from the additional channel in this pipeline. Figure~\ref{fig:bev_exp2_exp5} visualizes this effect: Exp~5 produces denser foreground clusters but removes much of the background/ground structure used for stable localization.
\paragraph{Exp~0 (probe baseline)}
Exp~0 uses oracle (ground-truth) 2D boxes to define frustums and fits 3D boxes by clustering and PCA/SVD, serving as a non-learning
depth-quality probe rather than a deployable detector. While it attains small non-zero AP at IoU$=0.5$, it collapses under the strict
IoU$=0.7$ criterion (0.28 AP$_{3D}$ on Moderate), indicating that heuristic frustum box fitting is insufficient for reliable 3D localization even with perfect 2D boxes.

\paragraph{Why BEV can remain high while 3D drops}
BEV IoU evaluates overlap only in the ground plane and ignores vertical alignment and height extent, whereas 3D IoU
penalizes errors in object height and vertical placement. This explains cases such as Exp~4, where AP$_{BEV}$ at IoU$=0.7$
on Moderate remains competitive (11.36), but AP$_{3D}$ drops sharply (5.73). This pattern is consistent with boxes that
are approximately correct in $x$--$y$ but misestimated in height or vertical position, which is especially costly at IoU$=0.7$.

\paragraph{Semantic channels and context reduction}
Comparing Exp~2 (grayscale) with Exp~7 (zero-intensity control) shows only a small difference
(19.06 vs.\ 18.53 AP$_{3D}$ at IoU$=0.5$; 10.50 vs.\ 9.83 at IoU$=0.7$), indicating that the 4th channel provides only
marginal gains relative to geometry. Mask-confidence injection (Exp~4) does not improve strict 3D overlap, and mask-guided
point selection (Exp~5) can reduce performance by removing background/ground context that supports stable localization
(Figure~\ref{fig:bev_exp2_exp5}).

\subsection{Depth Backbone Swap (Exp~2 feature setting)}
Table~\ref{tab:depth_model_comp} isolates the impact of the depth backbone by holding the detector, training protocol,
and pseudo-LiDAR feature setting fixed while swapping only the depth model. NeWCRFs slightly outperforms Depth Anything V2 Metric-Outdoor (Base)
across both IoU settings, with the gap more pronounced under the strict IoU$=0.7$ criterion (Moderate AP$_{3D}$:
10.50\% vs.\ 9.79\%). This supports the conclusion that improved geometric fidelity in the reconstructed point cloud leads to tighter 3D box alignment. Figure~\ref{fig:depth_newcrfs_da} compares the predicted depth maps, and Figure~\ref{fig:bev_newcrfs_da} shows the corresponding downstream BEV behavior under the same Exp~2 pipeline.

To better interpret how depth quality varies with range, we next report depth accuracy versus distance using ground-truth 2D boxes (Table~\ref{tab:depth_range_acc}).

\subsection{Depth Accuracy vs.\ Distance (GT 2D Boxes)}
We emphasize that 3D detection experiments in this paper are \textbf{Car-only}; Ped/Cyc are included here only for the depth diagnostic.

Table~\ref{tab:depth_range_acc} reports a coarse depth diagnostic using ground-truth 2D boxes (correct if
$|d_{\text{pred}}-d_{\text{gt}}|\le 1.5$\,m). We define $d_{\text{pred}}$ as the median predicted depth inside the 2D box
(after validity filtering $1 < D(u,v) < 60$\,m) and $d_{\text{gt}}$ as KITTI \texttt{location.z} (camera coordinates).
While this metric captures range consistency, it does not measure boundary sharpness or vertical alignment, which dominate strict AP$_{3D}$ at IoU$=0.7$.

\begin{table}[t]
\centering
\caption{Depth accuracy versus distance using ground-truth 2D boxes (correct if $|d_{\text{pred}}-d_{\text{gt}}|\le 1.5$\,m). Values are accuracy (\%) per class.}
\label{tab:depth_range_acc}
\scriptsize
\setlength{\tabcolsep}{4.2pt}
\renewcommand{\arraystretch}{1.08}
\begin{tabular}{l c c c c}
\toprule
Model & Range (m) & Car & Ped & Cyc \\
\midrule
\multirow{3}{*}{NeWCRFs~\cite{Yuan_2022_CVPR}}
& $\le 20$ & \textbf{60.6} & \textbf{67.7} & \textbf{54.0} \\
& $\le 40$ & 38.6 & \textbf{59.1} & \textbf{43.3} \\
& $\le 80$ & 33.3 & \textbf{56.7} & \textbf{36.8} \\
\midrule
\multirow{3}{*}{Depth Anything V2 Metric-Outdoor (Base)~\cite{depth_anything_v2}}
& $\le 20$ & 59.8 & 18.9 & 24.3 \\
& $\le 40$ & \textbf{38.7} & 17.4 & 23.8 \\
& $\le 80$ & \textbf{33.4} & 17.1 & 19.3 \\

\bottomrule
\end{tabular}
\end{table}
\begin{table*}[!t]
\centering
\caption{Comparison with representative monocular methods on KITTI for \textbf{Car}. Our method is evaluated on the KITTI validation split (train: 3{,}712 / val: 3{,}769) under AP$_{40}$. Other rows report numbers copied from the cited papers/releases.}

\label{tab:sota_comp_val_joined}

\scriptsize
\setlength{\tabcolsep}{2.0pt}
\renewcommand{\arraystretch}{1.10}

\begin{tabular*}{\textwidth}{@{\extracolsep{\fill}} l ccc ccc}
\toprule
& \multicolumn{3}{c}{AP$_{BEV}$/AP$_{3D}$ (\%), IoU=0.5} & \multicolumn{3}{c}{AP$_{BEV}$/AP$_{3D}$ (\%), IoU=0.7} \\
\cmidrule(lr){2-4}\cmidrule(lr){5-7}
Method & Easy & Moderate & Hard & Easy & Moderate & Hard \\
\midrule

\textbf{Ours (Exp~2, NeWCRFs~\cite{Yuan_2022_CVPR} + PointRCNN~\cite{Shi_2019_CVPR})} &
29.72/26.90 & 22.60/19.06 & 19.36/17.97 &
13.85/11.88 & 11.72/10.50 & 11.20/9.98 \\

\midrule

CenterNet~\cite{Zhou_2019_CenterNet}$^{\ast}$ &
34.36/20.00 & 27.91/17.50 & 24.65/15.57 &
3.46/0.60 & 3.31/0.66 & 3.21/0.77 \\

MonoGRNet~\cite{Qin2019MonoGRNet}$^{\ast}$ &
52.13/47.59 & 35.99/32.28 & 28.72/25.50 &
19.72/11.90 & 12.81/7.56 & 10.15/5.76 \\

M3D-RPN~\cite{Brazil_2019_M3DRPN}$^{\ast}$ &
53.35/48.53 & 39.60/35.94 & 31.76/28.59 &
20.85/14.53 & 15.62/11.07 & 11.88/8.65 \\

MonoPair~\cite{chen2020cvpr} &
61.06/55.38 & 47.63/42.39 & 41.92/37.99 &
24.12/16.28 & 18.17/12.30 & 15.76/10.42 \\

\midrule

MonoDLE~\cite{Ma_2021_MonoDLE} &
60.73/55.41 & 46.87/43.42 & 41.89/37.81 &
24.97/17.45 & 19.33/13.66 & 17.01/11.68 \\

Kinematic3D~\cite{Brazil_2020_Kinematic3D} &
61.79/55.44 & 44.68/39.47 & 34.56/31.26 &
27.83/19.76 & 19.72/14.10 & 15.10/10.47 \\

GrooMeD-NMS~\cite{Kumar_2021_GrooMeDNMS} &
61.83/55.62 & 44.98/41.07 & 36.29/32.89 &
27.38/19.67 & 19.75/14.32 & 15.92/11.27 \\

GUPNet~\cite{Lu_2021_GUPNet} &
61.78/57.62 & 47.06/42.33 & 40.88/37.59 &
31.07/22.76 & 22.94/16.46 & 19.75/13.72 \\

MonoDTR~\cite{Huang_2022_MonoDTR} &
69.04/64.03 & 52.47/47.32 & 45.90/42.20 &
33.33/24.52 & 25.35/18.57 & 21.68/15.51 \\

\midrule

MonoDIS~\cite{Simonelli_2019_MonoDIS} &
-- & -- & -- &
18.45/11.06 & 12.58/7.60 & 10.66/6.37 \\

MonoRUn~\cite{Chen_2021_MonoRUn} &
--/59.71 & --/43.39 & --/38.44 &
--/20.02 & --/14.65 & --/12.61 \\

CaDDN~\cite{CaDDN} &
-- & -- & -- &
--/23.57 & --/16.31 & --/13.84 \\

\bottomrule
\end{tabular*}

\vspace{0.25em}
\footnotesize{\textbf{Note:} $^{\ast}$ indicates values copied from the cited papers/releases.}
\end{table*}

\subsection{Comparison with Representative Monocular Methods (KITTI validation)}
\label{subsec:sota}

Table~\ref{tab:sota_comp_val_joined} compares our pseudo-LiDAR + PointRCNN pipeline (NeWCRFs, Exp~2) against
representative \emph{monocular} 3D car detectors as reported in the cited papers/releases (AP$_{40}$ where available).

Overall, our strict IoU$=0.7$ result (10.50 AP$_{3D}$ on Moderate) exceeds early/weak monocular baselines
(e.g., CenterNet and MonoDIS), is comparable to some mid-era approaches (e.g., M3D-RPN),
but remains below modern end-to-end monocular detectors on the same split (e.g., MonoDTR).
This indicates that pseudo-LiDAR + an off-the-shelf LiDAR detector is viable, but dedicated monocular 3D
architectures still retain a clear advantage.

\section{Discussion}
\label{sec:discussion}

Our experiments suggest three takeaways. \textbf{(i) Depth backbone dominates}: swapping only the depth model
changes strict AP$_{3D}$@0.7 more than any 4th-channel feature (Table~\ref{tab:depth_model_comp}).
\textbf{(ii) Semantic injection is not free}: using mask-confidence as the intensity channel can preserve BEV overlap
yet reduce AP$_{3D}$ at IoU$=0.7$ (Table~\ref{tab:car_apbev_ap3d}), consistent with sensitivity to height/vertical placement and the fact that PointRCNN was designed around LiDAR-style reflectance statistics rather than semantic confidence.
\textbf{(iii) Context matters}: mask-guided sampling (Exp~5) reduces ground/background structure that PointRCNN implicitly uses for stable localization, which can harm strict 3D overlap; however Exp~5 also increases the point budget (40k), so the result reflects a combined effect of sampling and density.

\paragraph{Limitations and future work}
We use a fixed LiDAR detector (PointRCNN) and simple feature injection without explicitly adapting the architecture
to semantic cues. Future work includes (a) modifying LiDAR backbones to consume semantic channels (e.g., dedicated
fusion branches or feature-wise conditioning), and (b) training or fine-tuning the depth model with 3D supervision
(e.g., box-consistent depth or detection-aware depth losses) to better align reconstructed geometry with 3D IoU.

\section{Conclusion}
\label{sec:conclusion}

We presented a systematic study of how depth backbones and semantic feature engineering affect monocular 3D object detection in a pseudo-LiDAR pipeline on KITTI. By holding the detector architecture and training protocol fixed, we isolate the contribution of the input reconstruction and feature channels.

Our main finding is that \textbf{depth-backbone choice dominates performance}. Under an identical pseudo-LiDAR feature setting and PointRCNN training protocol, \textbf{NeWCRFs} consistently but slightly outperforms \textbf{Depth Anything V2 Metric-Outdoor (Base)}, with the most meaningful gains under the strict IoU$=0.7$ criterion (10.50\% AP$_{3D}$ on Moderate). This reflects the fact that strict 3D IoU is highly sensitive to structured metric errors in the reconstructed geometry, including depth bias and vertical misalignment.

Across semantic feature variants, we find that \textbf{geometry remains the dominant signal} for PointRCNN in the pseudo-LiDAR setting: grayscale intensity yields only modest gains over a zero-intensity control, mask-confidence injection does not reliably improve strict 3D overlap, and mask-guided point selection can reduce accuracy by removing essential contextual structure. These outcomes indicate that \textbf{semantic feature injection alone is not sufficient} to substantially improve performance when using an off-the-shelf LiDAR detector; extracting value from semantics likely requires explicit fusion-aware architectural changes.

Taken together, our results suggest that improving monocular 3D detection through pseudo-LiDAR is primarily a problem of producing geometrically faithful reconstructions and matching the detector architecture to the statistics of the provided channels.



\bibliographystyle{IEEEtran}
\bibliography{refs}

\end{document}